\documentclass[10pt,twocolumn,letterpaper]{article}

\usepackage{cvpr}
\usepackage{cuted}
\usepackage{currfile}
\usepackage{caption}
\usepackage{booktabs}
\usepackage{multirow}
\usepackage{amsmath,amssymb}
\usepackage{bm}
\usepackage{graphicx}
\usepackage{float}
\usepackage{xspace}
\usepackage{enumitem}
\setlist{nosep,leftmargin=*}

\newcommand{\TODO}[1]{}

\newcommand{\model}{World Dreamer v5-lite\xspace}

\newcommand{\phistate}{\pmb{\phi}}

\usepackage[hidelinks]{hyperref}

\def\confName{CVPR}
\def\confYear{2027}

\title{\model: A Multi-Robot World Action Model\\with One Dreamer}

\author{
Guoning Wu\thanks{Guoning Wu and Zijian Cai contributed equally.} \quad Zijian Cai\footnotemark[1] \quad Yuhang Zhang \\
Ying Liu \quad Yongbin Zheng\thanks{Corresponding author.} \\
World Agents
}

\begin{document}
\raggedbottom
\maketitle
\begin{abstract}
Multi-robot control requires integrating distributed camera views while predicting separate action sequences.
Most vision--language--action and world-action models target one robot.
Combining multiple views alone does not establish spatial correspondence between them.
\model is a multi-robot world-action model built around a shared world state $\phistate$ for robot-specific action prediction.
Calibrated features from a pretrained video encoder populate $\phistate$ in workspace coordinates; its spatial size remains fixed as views are added.
A vision--language model provides scene context and one query per robot.
The queries read $\phistate$ and exchange pose-aware information, while their action streams remain separate.
Masked-view cooperative training withholds selected camera views from both visual branches while retaining all action targets.
This encourages the use of observations shared by other robots.
Pretraining uses more than $10{,}000$ hours of trajectories from 20 simulated robot types, RoboFactory, and real robots.
On RoboFactory, \model achieves $74.0\%$ mean success, 23.7 percentage points above the strongest evaluated baseline.
Across four real-robot tasks, it reaches $78.8\%$ mean success, a margin of 13.8 points.
It also reaches $58.08\%$ overall on RoboCasa365.
\end{abstract}

\section{Introduction}
\label{sec:intro}

Vision--language--action (VLA) models adapt pretrained multimodal backbones to robot control~\cite{brohan2023rt2,kim2024openvla,black2024pi0,black2025pi05,bjorck2025groot}, and world-action models further couple action prediction with predicted visual dynamics~\cite{du2023universal,li2025uva,zhu2025uwm}.
These models inherit semantic knowledge from vision--language pretraining and, in the case of world-action models, knowledge of how scenes evolve from video pretraining.
Most of these models predict actions for a single robot.

\begin{figure*}[t]
  \centering
  \includegraphics[width=\textwidth]{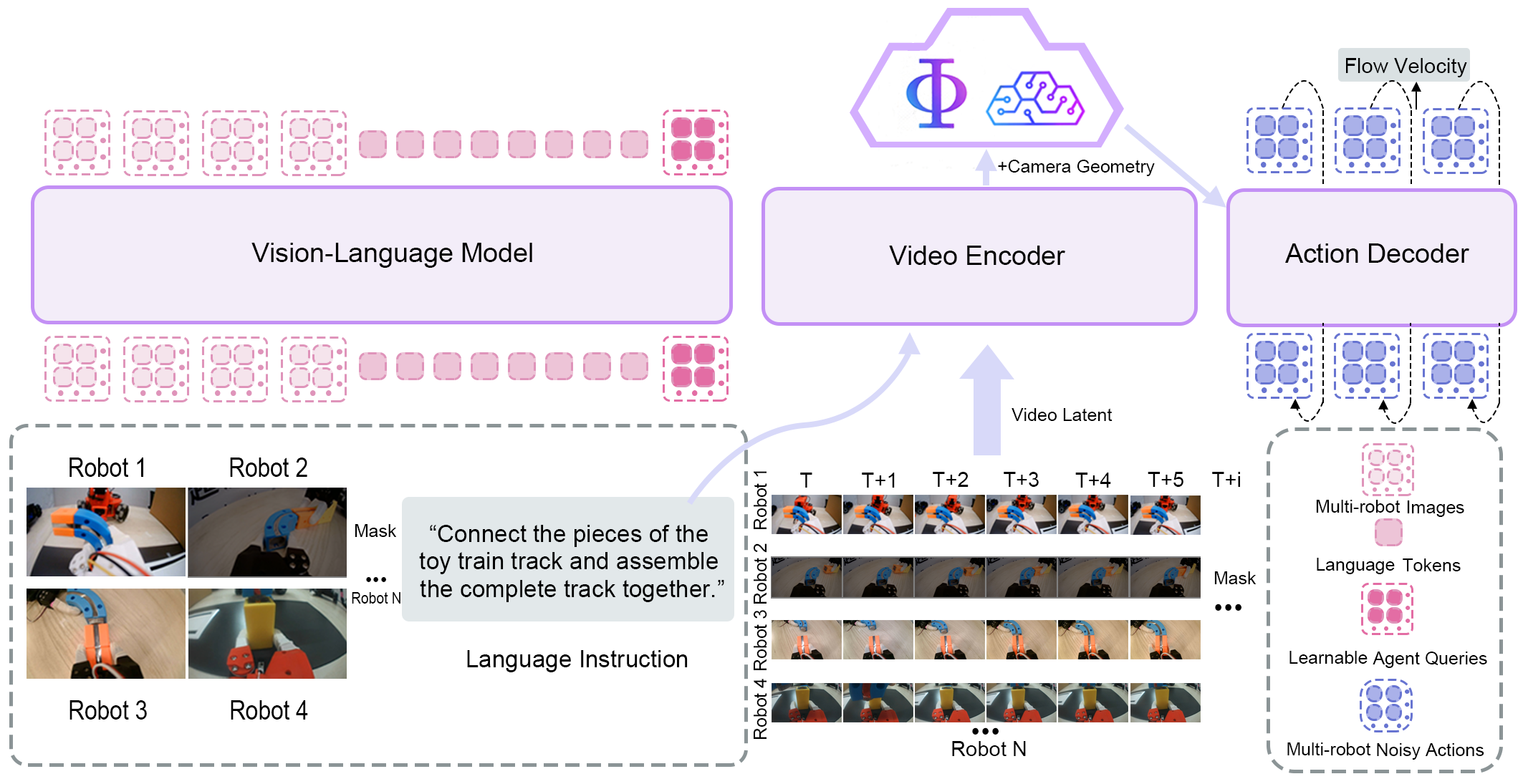}
  \caption{\textbf{Overall architecture of \model.}
  Multi-robot images and language tokens enter the vision--language model (VLM) before the learnable agent queries, yielding a scene-prefix key/value (KV) cache and one hidden state per robot query.
  The video strip depicts a temporal window: observed video latents are encoded and combined with camera geometry to build the shared world state $\phistate$; future latents supply training supervision only.
  The Action Decoder uses the VLM outputs, $\phistate$, proprioception, and multi-robot noisy actions to predict a flow velocity for each robot.}
  \label{fig:arch}
\end{figure*}

Many manipulation tasks instead require several robots to work in a shared workspace, for example, when one robot holds an object while another operates on it~\cite{qin2025robofactory}.
This setting differs from single-robot control in three respects.
First, observations are distributed: each robot typically carries its own cameras, which move with it and capture only part of the workspace, and image patches from different cameras do not correspond to the same physical locations.
Second, control is separate while decisions are coupled: the robots may differ in embodiment and action space, yet the action of each robot depends on where the other robots are and what they are doing.
Third, observations are often incomplete: the camera of one robot may be occluded by another robot or by the manipulated object, so a robot must sometimes act on what other cameras observe.

Extending single-robot models to this setting is not straightforward.
Concatenating camera tokens does not, by itself, establish correspondence between views in workspace coordinates.
Separate policies preserve robot-specific controls, but a policy restricted to its own cameras cannot use observations from teammates.
Existing multi-agent policies can share information across robots~\cite{wang2025gaudp,park2026clsdp}.
Even with this capacity, training with complete observations gives a robot little incentive to use another view when its own cameras already suffice.

\model addresses these requirements with a shared world state $\phistate$ built from calibrated video features in workspace coordinates.
A vision--language model conditions one query per robot on the current images and instruction; the queries read $\phistate$ and exchange pose-aware information while their action streams remain separate.
Only observed video features enter the state used for action prediction, while an auxiliary future-video latent objective trains the video encoder.
During training, selected camera views are withheld from both visual branches, but every robot retains its action target, exposing the model to incomplete visual coverage.

Evaluation spans the multi-robot benchmark RoboFactory~\cite{qin2025robofactory}\footnote{The camera configuration of RoboFactory is modified so that each agent receives observations from its own cameras. All methods are evaluated under this setting.}, the household benchmark RoboCasa365~\cite{nasiriany2024robocasa,nasiriany2026robocasa365}, and four real-robot tasks.
The real-robot tasks use homogeneous LeKiwi teams~\cite{lekiwi2025} and heterogeneous LeKiwi--OpenArm teams~\cite{openarm2025}.

The main contributions of this work are summarized as follows:
\begin{itemize}[leftmargin=*]
  \item A multi-robot manipulation framework, \model, is proposed. Calibrated multi-view features from a pretrained video encoder form a shared world state $\phistate$ read by all robots. Pose-aware agent queries exchange information across robots.
  \item A masked-view cooperative training strategy is introduced. It withholds robot views from both visual branches while retaining all action supervision. This encourages each robot to act on observations shared by the others.
  \item \model achieves leading performance among the evaluated methods on RoboFactory and four real-robot tasks. Its mean success rate exceeds the strongest baselines by 23.7 and 13.8 percentage points, respectively. On RoboCasa365, it achieves competitive performance.
\end{itemize}

\section{Related Work}
\label{sec:related}

\paragraph{Vision--language--action models.}
Vision--language--action (VLA) policies map language and images to robot actions with pretrained multimodal backbones.
RT-2 co-fine-tunes web-pretrained vision--language models (VLMs) on robot trajectories and represents actions as text tokens~\cite{brohan2023rt2}.
OpenVLA builds on the Prismatic VLM and predicts discretized action tokens after training on about 970k demonstrations from Open~X-Embodiment~\cite{kim2024openvla}, and Octo is a transformer policy pretrained on 800k Open~X-Embodiment trajectories and instructed by language or goal images~\cite{team2024octo}.
$\pi_{0}$ attaches a flow-matching action expert to a PaliGemma backbone~\cite{black2024pi0}, and $\pi_{0.5}$ extends its training to heterogeneous data sources for open-world mobile manipulation~\cite{black2025pi05}.
GR00T~N1 pairs a vision--language module with a diffusion transformer that generates continuous actions~\cite{bjorck2025groot}.
These policies process the observations of one robot as a sequence of image tokens.
\model retains the VLM as a source of semantic context and agent queries, but additionally aggregates the views of all robots in a calibrated shared world state.

\paragraph{World-action models.}
World-action models couple action prediction with predicted visual dynamics.
UniPi generates a future video from an image and a language goal and extracts actions with an inverse-dynamics model~\cite{du2023universal}.
UWM integrates action diffusion and video diffusion in one transformer with an independent diffusion timestep for each modality~\cite{zhu2025uwm}.
Several recent models use video prediction mainly as a training signal: UVA decodes video and actions from a joint latent with separate diffusion heads and bypasses video generation at inference~\cite{li2025uva}, and Fast-WAM~\cite{yuan2026fastwam} and GigaWorld-Policy~\cite{ye2026gigaworldpolicy} likewise retain video co-training while skipping future-frame generation when acting.
\model also uses future-frame prediction only during training, but it lifts the observed features of the video encoder into a shared world state in workspace coordinates instead of using them as image-indexed tokens of a single robot.

\paragraph{Multi-robot policies.}
RoboFactory provides compositional multi-agent manipulation tasks for studying embodied collaboration~\cite{qin2025robofactory}.
On this benchmark, GauDP reconstructs a 3D Gaussian field from the RGB observations of all agents and redistributes it to each agent within a diffusion policy~\cite{wang2025gaudp}.
CLS-DP instead targets decentralized execution without communication: privileged multi-agent dynamics are distilled into a latent space during training, and each agent infers this latent from its own camera at deployment~\cite{park2026clsdp}.
\model differs from these methods in three respects: it builds on pretrained vision--language and video models, it lets robots exchange information explicitly through pose-aware agent queries, and its masked-view training encourages robots to act on shared observations when their own views are withheld.

\paragraph{3D visuomotor policies.}
Explicit 3D representations provide spatial structure for visuomotor policies.
DP3 conditions a diffusion policy on a compact representation extracted from sparse point clouds~\cite{ze2024dp3}.
PDFactor factorizes a point cloud into three orthogonal feature planes~\cite{tian2025pdfactor}, a plane factorization also used for 3D-aware generation in EG3D~\cite{chan2022eg3d}.
The shared world state $\phistate$ adopts a similar plane layout but is built from the features of a pretrained video encoder across multiple moving cameras, handles patches without depth through candidate points along their viewing rays, and is accumulated over observed frames with a visibility-gated update.

\section{Method}
\label{sec:method}

In this section, we present \model, a framework for language-conditioned control of multiple robots that share a workspace.
At each decision step, given a language instruction, the current and recent RGB observations of calibrated cameras, and the proprioception of each robot, \model predicts an action chunk for every robot.
As illustrated in Fig.~\ref{fig:arch}, \model draws shared context from two sources.
A vision--language model (VLM) processes the instruction and the current images together with one learnable query per robot (Sec.~\ref{sec:vlm}).
In parallel, a video encoder processes the recent observations of all cameras (Sec.~\ref{sec:video}), and its features are placed in workspace coordinates to build the shared world state $\phistate$ (Sec.~\ref{sec:worldstate}).
The Action Decoder combines these signals with the proprioception of each robot and the relative poses between robots to predict the controls of every robot (Sec.~\ref{sec:action}).
Finally, Sec.~\ref{sec:training-objective} introduces masked-view cooperative training, which encourages the robots to act on each other's observations.

\begin{figure*}[t]
  \centering
  \includegraphics[width=\textwidth]{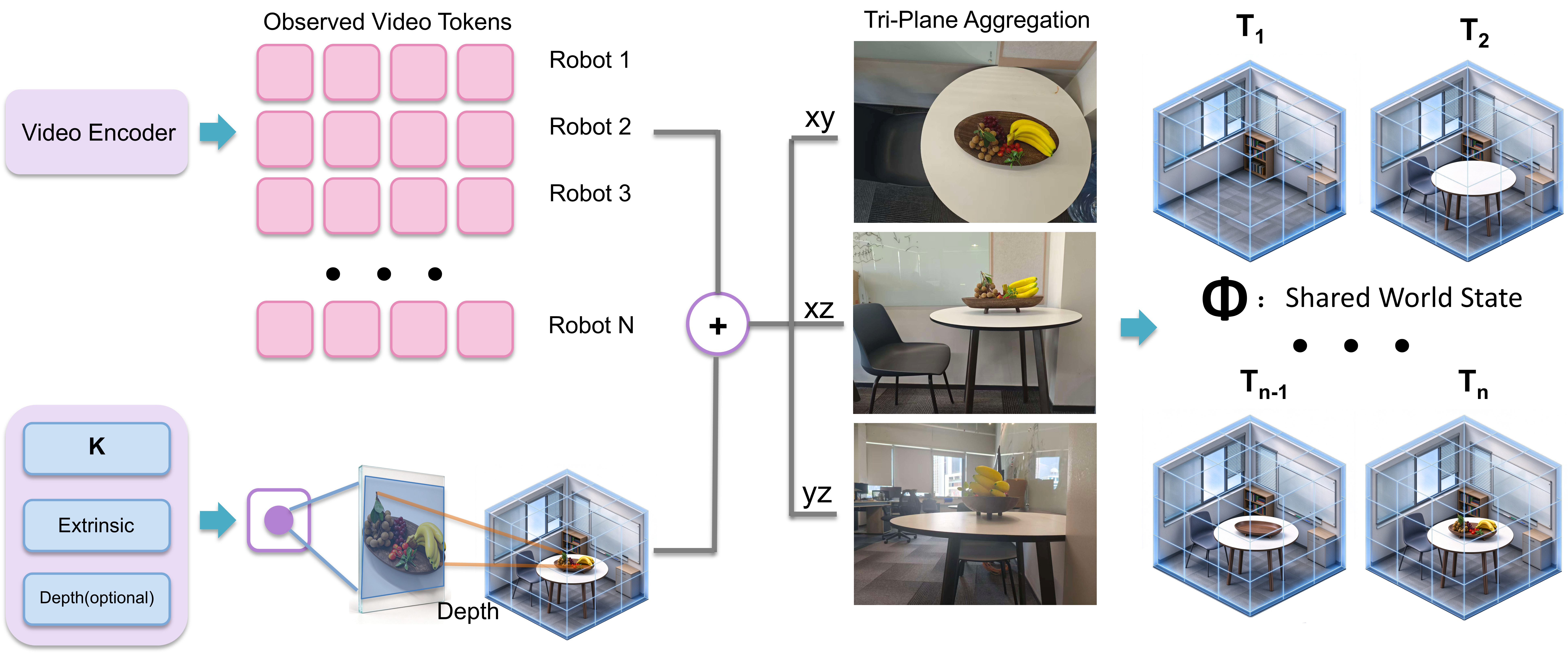}
  \caption{\textbf{Constructing the shared world state $\phistate$.}
  Observed video tokens from available views are placed in 3D using the intrinsics $K$, the extrinsics, and optional depth.
  Each token is fused with its geometric embedding ($+$) and written to the XY, XZ, and YZ planes.
  The right-hand panels schematically show how $\phistate$ is updated over successive observed frames.}
  \label{fig:worldstate}
\end{figure*}

\subsection{Vision--language conditioning}
\label{sec:vlm}

The instruction and the current images of the available views form a shared prefix of language and image tokens for the VLM (Fig.~\ref{fig:arch}).
A fixed layout reserves $N$ robot slots, and an agent mask $\mathbf{m}^{\mathrm{agent}}\in\{0,1\}^N$ selects the robots present at the decision step.
One learnable agent query per robot slot follows the prefix.
The input embedding of the query for slot $i\in\{1,\ldots,N\}$ is
\begin{equation}
\mathbf{r}_i=\mathbf{q}_{\mathrm{shared}}+\mathbf{e}_i,
\end{equation}
where $\mathbf{q}_{\mathrm{shared}}$ is a learnable embedding common to all robots and $\mathbf{e}_i$ is a learnable embedding of slot $i$.
Because all queries follow the same prefix, one VLM pass conditions the robot queries on the same scene without rerunning the backbone separately for each robot.

The VLM provides two outputs to the Action Decoder.
First, the VLM processes the input query embedding $\mathbf{r}_i$ with the shared scene prefix and produces a contextualized hidden state at robot query position $i$.
This hidden state is projected and normalized to form the initial decoder query $\mathbf{q}_i^{0}\in\mathbb{R}^{d}$, where $d{=}1024$ is the decoder width.
Second, the layer-wise keys and values of the prefix tokens form a shared scene-prefix cache, which supplies the instruction and image context to every robot without another VLM pass (Sec.~\ref{sec:action}).
Outputs of unused slots are zero-padded.

\subsection{Video encoder}
\label{sec:video}

While the VLM provides semantic context from the current images, the video encoder supplies the spatial evidence from which $\phistate$ is built.
Recent RGB histories occupy $V$ camera slots, which are separate from the $N$ robot slots: a robot may provide several views, and a workspace camera may provide another.
A view mask $\mathbf{m}^{\mathrm{view}}\in\{0,1\}^V$ selects the available cameras; the same mask determines which current images enter the VLM.

A causal video variational autoencoder (VAE) converts each RGB history into latents, and the encoder of a pretrained video model processes $T_o{=}3$ observed latent steps.
During training, $T_f{=}2$ noised future steps are additionally provided for an auxiliary video-flow prediction task.
A prefix-asymmetric attention mask prevents observed tokens from attending to the future steps.
The observed outputs, and therefore $\phistate$, depend only on observations, so no video generation is required at inference.

\subsection{Shared world state \texorpdfstring{$\phistate$}{phi}}
\label{sec:worldstate}

The video encoder produces a feature for each observed image patch.
Patch indices from different cameras do not identify the same physical location, so features from different views cannot be pooled directly, and concatenating them would yield a token set that grows with the number of cameras.
The shared world state $\phistate$ instead organizes features by their locations in a common workspace.
As shown in Fig.~\ref{fig:worldstate}, camera geometry places each token in 3D, the token is fused with its geometric embedding, and the result is written to three axis-aligned planes (XY, XZ, and YZ)~\cite{chan2022eg3d,tian2025pdfactor}, which are updated over the observed frames.
The resulting state has a fixed size regardless of the number of cameras and is shared by all robots.

\paragraph{From image patches to 3D locations.}
For view $v$ at observed time $t$, let $K_{vt}$ be the intrinsic matrix and $R_{vt},\mathbf{t}_{vt}$ the world-to-camera rotation and translation.
For patch index $p$ with homogeneous center pixel $\tilde{\mathbf{p}}$, the camera origin $\mathbf{o}_{vt}$ and the unit ray $\mathbf{d}_{vtp}$ are
\begin{equation}
\mathbf{o}_{vt}=-R_{vt}^{\top}\mathbf{t}_{vt},\quad
\mathbf{d}_{vtp}=\operatorname{norm}\!\left(R_{vt}^{\top}K_{vt}^{-1}\tilde{\mathbf{p}}\right),
\end{equation}
where $\operatorname{norm}$ rescales a vector to unit length.
When a valid depth $D_{vtp}$ along the optical axis is available, the patch maps to the point $\mathbf{o}_{vt}+D_{vtp}R_{vt}^{\top}K_{vt}^{-1}\tilde{\mathbf{p}}$.
Without depth, a patch specifies a ray rather than a unique 3D point.
It is then represented by $S{=}8$ candidate points along the ray within the workspace, whose contributions are learned from their features and geometry.
Samples outside the workspace and samples from views with $\mathbf{m}^{\mathrm{view}}_v{=}0$ do not contribute.
During training, depth is dropped with probability $0.25$, so that the same model handles both measured points and depth-free rays.

\paragraph{Building $\phistate$.}
Let $k=(v,p,s)$ index a valid sample from view $v$, patch $p$, and ray position $s\in\{1,\ldots,S\}$.
Its fused feature $\mathbf{z}_k$ is obtained by adding an embedding of its geometry to the projected token, which corresponds to the plus sign in Fig.~\ref{fig:worldstate}.
The sample also has a ray weight $w_k$ and a location $\mathbf{x}_k$ in normalized workspace coordinates.
With depth, $w_k$ selects the measured point; without depth, $w_k$ is learned over the candidates along the ray.
For plane $ab\in\{XY,XZ,YZ\}$ and grid location $u$, the samples at observed time $t$ are written by weighted bilinear splatting:
\begin{equation}
\begin{aligned}
\alpha^{ab}_{k,u}&=w_k\,\kappa\!\left(u,\pi_{ab}(\mathbf{x}_k)\right),\\
\mathbf{M}^{ab}_t(u)&=\frac{\sum_k\alpha^{ab}_{k,u}\mathbf{z}_k}
 {\max\!\left(\delta,\sum_k\alpha^{ab}_{k,u}\right)},
\end{aligned}
\end{equation}
where $\pi_{ab}$ projects a 3D point onto plane $ab$, $\kappa$ is the bilinear kernel, and $\delta$ is a small constant.
Normalizing by the accumulated weight keeps the feature scale independent of the number of views that observe a location.
The aggregated planes are then mixed by lightweight MLPs to produce the candidate state $\tilde{\phistate}_t$.

\paragraph{Visibility-gated update.}
Cameras mounted on the robots move, and each frame observes only part of the workspace.
Replacing the state at every step would therefore erase regions that are temporarily out of view.
Instead, starting from $\phistate_{0}{=}\mathbf{0}$, the state is updated for $t{=}1,\ldots,T_o$ as
\begin{equation}
\phistate_{t}
=
\phistate_{t-1}
+
\mathbf{g}_{t}
\odot
\bigl(\tilde{\phistate}_{t}-\phistate_{t-1}\bigr),
\qquad
\mathbf{g}_{t}
=
\sigma\!\bigl(\mathrm{MLP}(\mathbf{h}_{t})\bigr)\odot\mathbf{v}_{t},
\end{equation}
where $\sigma$ is the sigmoid function and $\odot$ denotes elementwise multiplication.
The gate input $\mathbf{h}_{t}$ contains the previous and candidate features, and $\mathbf{v}_{t}\in[0,1]$ is the visibility of each grid location in the current frame.
The learned gate controls how much new evidence replaces the stored feature, and its product with $\mathbf{v}_{t}$ leaves locations unobserved in the current frame unchanged.
The Action Decoder receives the final state $\phistate=\phistate_{T_o}$, with $G{=}16$ and $C_f{=}256$:
\begin{equation}
\phistate\in\mathbb{R}^{3\times G\times G\times C_f}.
\end{equation}
This layout provides $3G^2$ locations for decoder readout, rather than $G^3$ in a dense 3D grid of the same resolution.

\subsection{Action Decoder}
\label{sec:action}

The Action Decoder must generate controls for robots that may differ in embodiment while accounting for how they interact.
Concatenating the action tokens of all robots into one sequence entangles heterogeneous action spaces and lengthens the sequence with every added robot, whereas decoding each robot independently ignores the others.
We therefore keep a separate action stream for each robot and let the robots communicate only through their queries $\mathbf{q}_i$, initialized from the VLM as $\mathbf{q}_i^{0}$ (Sec.~\ref{sec:vlm}).

For robot $i$, the current proprioception is $\mathbf{s}_i\in\mathbb{R}^{D_s}$ and the target action chunk is $\mathbf{a}_i\in\mathbb{R}^{H\times D_a}$, where $H{=}30$ is the number of future control steps.
Proprioception and actions are padded to common widths $D_s{=}20$ and $D_a{=}18$, which lets one decoder handle embodiments with different numbers of control coordinates.
The action mask $\mathbf{m}^{\mathrm{act}}\in\{0,1\}^{N\times H\times D_a}$ identifies the valid control coordinates and recorded future steps of each robot.

\begin{figure}[t]
  \centering
  \includegraphics[width=\columnwidth]{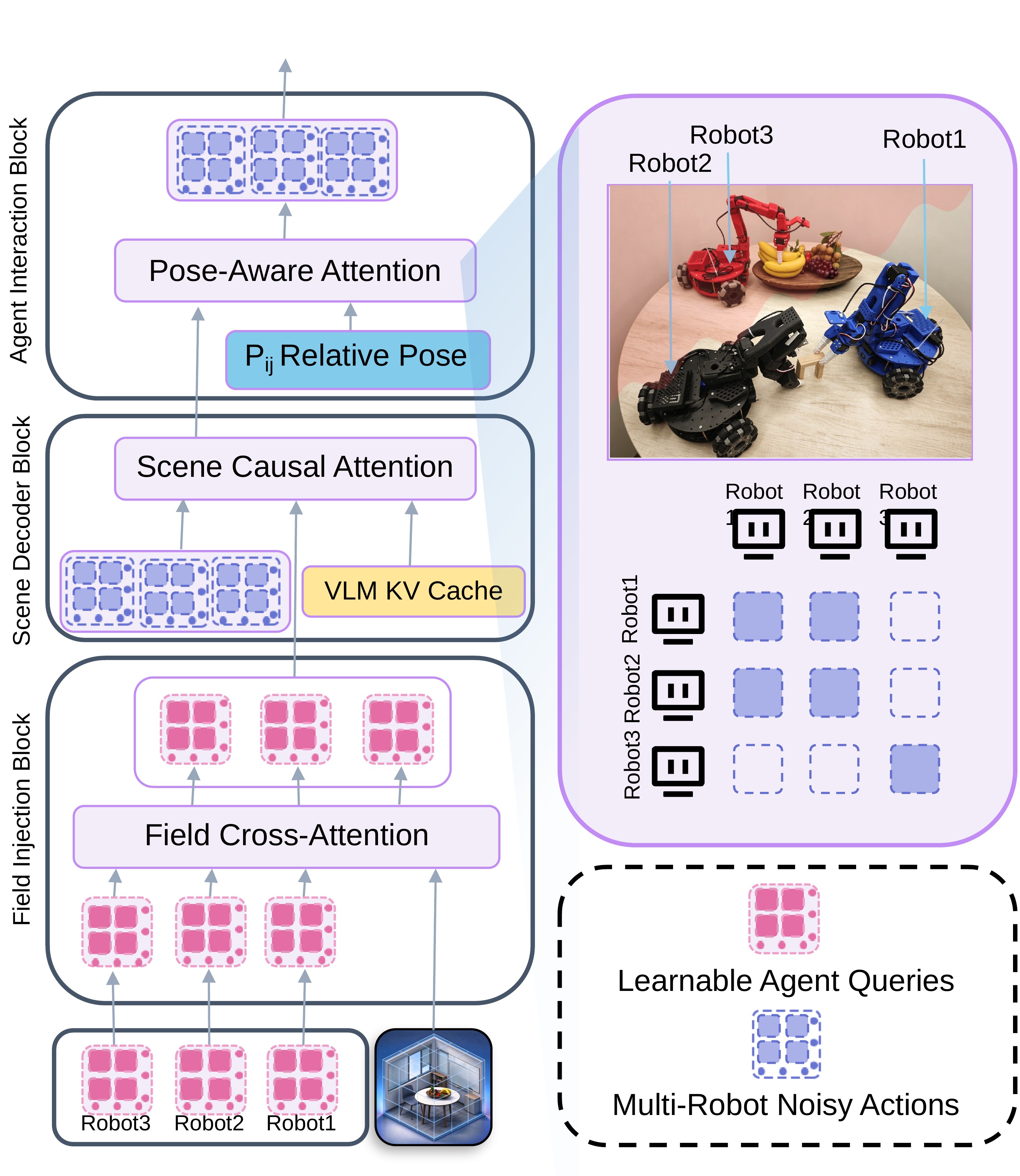}
  \caption{\textbf{Action Decoder information flow.}
  Pink denotes agent queries and blue denotes noisy action tokens.
  From bottom to top, the Field Injection Block reads $\phistate$ into each query, the Scene Decoder Block conditions the causal action stream of each robot on the VLM key/value (KV) cache, and the Agent Interaction Block exchanges queries through pose-aware attention.
  Right: the relative poses $\mathbf{P}_{ij}$ bias inter-robot attention. The action streams remain separate across robots.}
  \label{fig:query}
\end{figure}

The decoder is trained with conditional flow matching~\cite{liu2022rectified}.
Let $\boldsymbol{\epsilon}_i\sim\mathcal{N}(\mathbf{0},\mathbf{I})$ have the same shape as $\mathbf{a}_i$, and let $\tau\in[0,1]$ denote flow time.
The interpolated decoder input and its target velocity are
\begin{equation}
\tilde{\mathbf{a}}_i=(1-\tau)\boldsymbol{\epsilon}_i+\tau\mathbf{a}_i,
\qquad
\mathbf{u}^{\star}_i=\mathbf{a}_i-\boldsymbol{\epsilon}_i .
\end{equation}
The interpolated chunk $\tilde{\mathbf{a}}_i$ is projected into $H$ action tokens $\mathbf{A}_i\in\mathbb{R}^{H\times d}$.
Together with a learnable start embedding $\mathbf{e}_{\mathrm{start}}$, the projected proprioception $\bar{\mathbf{s}}_i$, and the query $\mathbf{q}_i$, they form the causal stream of robot $i$:
\begin{equation}
\bigl[\mathbf{e}_{\mathrm{start}},\ \bar{\mathbf{s}}_i,\ \mathbf{q}_i,\ \mathbf{A}_i\bigr].
\end{equation}
Each of the $L{=}36$ decoder layers updates the streams in the read--decode--communicate order of Fig.~\ref{fig:query}, so that the shared spatial evidence reaches the action tokens of a robot before its query is exchanged with the other robots.

First, the \textbf{Field Injection Block} lets each query $\mathbf{q}_i$ read $\phistate$ through cross-attention, retrieving the spatial context pooled from all cameras.
Because all robots read the same state, the attention is biased toward the location of the camera assigned to robot $i$, so that each robot reads $\phistate$ from its own perspective without excluding distant regions.

Second, the \textbf{Scene Decoder Block} conditions the action stream of robot $i$ on the scene-prefix cache.
Each token attends to the cache of the matching VLM layer and, causally, to the preceding tokens of its own stream.
An action token can therefore use the language--image context, the proprioception of its robot, the field-updated query, and the earlier action tokens, while remaining isolated from the action streams of the other robots.

Third, the \textbf{Agent Interaction Block} exchanges the queries of all robots through pose-aware attention.
The calibrated descriptor $\mathbf{P}_{ij}\in\mathbb{R}^{10}$ describes the camera assigned to robot $j$ in the frame of the camera assigned to robot $i$: 3-D translation, 6-D continuous rotation~\cite{zhou2019continuity}, and Euclidean distance.
A learned MLP maps it to the attention-logit bias $b_{ij}^{h}=f_h(\mathbf{P}_{ij})$ of head $h$.
As illustrated on the right of Fig.~\ref{fig:query}, these learned biases condition inter-robot exchange on relative pose without imposing a fixed connectivity pattern.
Because $\mathbf{P}_{ij}$ is relative, the bias does not depend on the choice of the world frame.
Only queries are exchanged, so information from other robots reaches the action tokens in the next layer, while each stream keeps its own proprioception and embodiment-specific action tokens.

An output head maps the final action tokens to the flow velocity $\hat{\mathbf{u}}_i\in\mathbb{R}^{H\times D_a}$.
At inference, $\mathbf{q}_i^{0}$, $\phistate$, and the scene-prefix cache are computed once per decision step; five Euler steps re-evaluate the decoder to integrate the predicted velocity from noise to an action chunk.

\subsection{Masked-view cooperative training}
\label{sec:training-objective}

The architecture above allows the robots to share information, but it does not by itself ensure that this information is used.
If every robot always observes its own cameras, the decoder can rely mainly on those views and ignore both $\phistate$ and the other robots.
To encourage cooperation under incomplete visual coverage, training uses $\mathbf{m}^{\mathrm{view}}$ to withhold either all cameras of one randomly selected robot or a random subset of views.
As illustrated in Fig.~\ref{fig:arch}, the same cameras are masked in both visual paths, so a masked camera contributes neither its current image to the VLM scene prefix nor its history to $\phistate$.
Camera calibration remains available to define spatial references and the relative-pose descriptors $\mathbf{P}_{ij}$.

Since view masking is applied only to observations, the recorded action chunks remain valid targets for all robots, including those whose cameras are hidden.
The remaining views can contribute through $\phistate$ and the scene-prefix cache, while query interaction provides another route for sharing context.

\paragraph{Training objective.}
Under the masked context, each robot retains its demonstrated action chunk as the target. The flow time is sampled as $\rho\sim\mathrm{Beta}(1.5,1)$ and $\tau=0.999(1-\rho)$, placing more samples near the noisy starting point. The action mask $\mathbf{m}^{\mathrm{act}}$ excludes unused controls and unrecorded steps when $\hat{\mathbf{u}}_i$ is compared with $\mathbf{u}^{\star}_i$.

The action loss $\mathcal{L}_{\mathrm{act}}$ is a weighted squared velocity error, averaged over the supervised entries of each robot and then over the robots present:
\begin{equation}
\mathcal{L}_{\mathrm{act}}=\mathbb{E}\Biggl[\frac{1}{|\mathcal{I}|}\sum_{i\in\mathcal{I}}\frac{1}{|\Omega_i|}\sum_{(h,c)\in\Omega_i}\omega_{ihc}\bigl(\hat{u}_{ihc}-u^{\star}_{ihc}\bigr)^{2}\Biggr],
\end{equation}
where the expectation is over scenes and flow times, $\mathcal{I}$ is the set of robots with supervised entries, $\Omega_i$ is the set of steps $h$ and control coordinates $c$ of robot $i$ selected by $\mathbf{m}^{\mathrm{act}}$, and $\omega_{ihc}$ is a per-entry weight.
With probability $0.5$, one to six demonstrated steps are supplied as an action prefix and only the remaining steps are supervised. For these examples, $\omega_{ihc}$ is the absolute action error of a detached rollout, normalized per robot and clipped to $[0.5,5]$; without a prefix, $\omega_{ihc}{=}1$.

Two auxiliary losses train the video and state-building paths. The generation tail predicts the noise-minus-latent velocity of the future video latents, and $\mathcal{L}_{\mathrm{future}}$ is the error of this prediction. For $\mathcal{L}_{\mathrm{field}}$, detached observed tokens pass through the state builder again. A rendering head samples the resulting state at the 3D locations of these tokens and predicts fixed random projections of the normalized observed tokens; the loss combines cosine distance with $0.1$ times Smooth-L1 error. Detachment confines its gradients to the builder and rendering head.

The full objective is
\begin{equation}
\mathcal{L}=0.5\,\mathcal{L}_{\mathrm{act}}+0.2\,\mathcal{L}_{\mathrm{future}}+\lambda_{\mathrm{field}}\mathcal{L}_{\mathrm{field}},
\end{equation}
where $\lambda_{\mathrm{field}}$ decreases linearly from $0.1$ to $0.02$ during training. Masking changes the visual inputs while preserving the action targets, allowing the remaining cameras to provide evidence for every robot's prediction.

\section{Experiments}
\label{sec:exp}

The evaluation covers multi-robot collaboration on RoboFactory~\cite{qin2025robofactory}, single-robot household manipulation on RoboCasa365~\cite{nasiriany2026robocasa365}, and real-robot manipulation with homogeneous and heterogeneous robot teams.

\subsection{Datasets and protocols}
\label{sec:data}

\paragraph{Pretraining corpus.}
The pretraining corpus combines manipulation trajectories generated in simulation for 20 robot types, RoboFactory trajectories, and real-robot trajectories collected with LeKiwi and OpenArm.
These sources cover single-arm and dual-arm configurations, more than 60 task types, and more than $10{,}000$ hours of interaction.
Differences in action dimensionality are handled by the action mask, and missing cameras are handled by the view mask.
The pretrained model is fine-tuned separately for each evaluation setting, and evaluation episodes are not included in the training data.

\paragraph{RoboFactory.}
RoboFactory provides compositional multi-agent manipulation tasks built on ManiSkill~\cite{qin2025robofactory}.
Evaluation uses three tasks with increasing numbers of agents: Place Food (two agents), Camera Alignment (three agents), and Take Photo (four agents).
These tasks are also evaluated in prior work on this benchmark~\cite{wang2025gaudp,park2026clsdp}, and they test how performance changes as the number of collaborating robots grows.
The original tasks are retained, but the camera configuration is modified so that each agent receives observations from its own cameras.
All methods in Tab.~\ref{tab:robofactory} are trained and evaluated under this configuration.
Following the RoboFactory protocol, each task is evaluated over 100 scene configurations.

\paragraph{RoboCasa365.}
RoboCasa365 provides 365 household tasks across 2,500 kitchen scenes, 220 of which require mobile manipulation~\cite{nasiriany2026robocasa365}.
Following the official leaderboard~\cite{robocasaLeaderboard2026}, \model is fine-tuned on the official Human300 pretraining dataset, which contains 300 tasks across 2,500 kitchen scenes; no other RoboCasa data are used.
Evaluation uses 50 tasks divided into three splits: 18 Atomic-Seen, 16 Composite-Seen, and 16 Composite-Unseen tasks.
The Composite-Unseen tasks are not included in Human300 and are therefore evaluated zero-shot.
Following the leaderboard protocol, each task is evaluated over 50 environments.

\paragraph{Real-robot tasks.}
Four real-robot tasks are evaluated: two are performed by two LeKiwi robots, and the other two by a heterogeneous team of two LeKiwi robots and one OpenArm.
Each trajectory records synchronized multi-view RGB images, proprioception, and camera calibration.
Each task is evaluated over 100 trials.

\subsection{Baselines and metrics}
\label{sec:baselines}

On RoboFactory and the real-robot tasks, \model is compared with six baselines: Diffusion Policy~\cite{chi2023diffusionpolicy}, ACT~\cite{zhao2023learning}, $\pi_{0.5}$~\cite{black2025pi05}, Fast-WAM~\cite{yuan2026fastwam}, and two variants of GauDP~\cite{wang2025gaudp}.
Global GauDP predicts the actions of all agents with a single joint policy, whereas Local GauDP controls each agent with an independent policy.
The remaining baselines are single-robot policies; one model is shared by all agents and is run once per agent at each decision step.
On RoboCasa365, \model is compared with recent advanced methods: PatchWAM~\cite{wang2026patchwam}, ABot-M0.5~\cite{chen2026abotm05}, PRTS~\cite{zhang2026prts}, RLDX-1~\cite{kim2026rldx1}, GigaWorld-Policy 0.1~\cite{ye2026gigaworldpolicy}, $\pi_{0.5}$~\cite{black2025pi05}, and Diffusion Policy~\cite{chi2023diffusionpolicy}.
Their results are taken from the official leaderboard~\cite{robocasaLeaderboard2026}.

The metric is the task success rate.
For RoboFactory and the real-robot tasks, the mean is the unweighted average over tasks.
For RoboCasa365, the overall score is the average over all 50 tasks, so the three splits are weighted by their numbers of tasks.

\subsection{Implementation details}
\label{sec:traindetails}

The vision--language backbone is initialized from Qwen3-VL~\cite{bai2025qwen3vl}.
Its vision encoder and language model are trained, while the input embeddings and the language-modeling head are frozen.
Training uses bfloat16 precision and DeepSpeed ZeRO Stage~2.
Each vision--language forward pass is reused for four flow-matching samples of the action decoder.

\subsection{Main results}
\label{sec:mainres}

\begin{table}[H]
\centering
\scriptsize
\setlength{\tabcolsep}{2pt}
\caption{Success rates (\%) on RoboFactory. All methods use the same camera configuration; Mean is the unweighted average over the three tasks.}
\label{tab:robofactory}
\begin{tabular}{@{}lcccc@{}}
\toprule
Method & \shortstack{Place\\Food} & \shortstack{Camera\\Alignment} & \shortstack{Take\\Photo} & Mean \\
\midrule
Diffusion Policy~\cite{chi2023diffusionpolicy} & 12 & 3 & 0 & 5.0 \\
Global GauDP~\cite{wang2025gaudp} & 18 & 28 & 5 & 17.0 \\
Local GauDP~\cite{wang2025gaudp} & 16 & 17 & 3 & 12.0 \\
$\pi_{0.5}$~\cite{black2025pi05} & 58 & 67 & 26 & 50.3 \\
ACT~\cite{zhao2023learning} & 13 & 21 & 10 & 14.7 \\
Fast-WAM~\cite{yuan2026fastwam} & 23 & 56 & 16 & 31.7 \\
\midrule
\textbf{Ours} & \textbf{94} & \textbf{78} & \textbf{50} & \textbf{74.0} \\
\bottomrule
\end{tabular}
\end{table}

\paragraph{RoboFactory.}
As shown in Tab.~\ref{tab:robofactory}, \model achieves a mean success rate of $74.0\%$, compared with $50.3\%$ for the strongest baseline, $\pi_{0.5}$.
It achieves the highest success rate on all three tasks, exceeding $\pi_{0.5}$ by 36, 11, and 24 percentage points on Place Food, Camera Alignment, and Take Photo, respectively.

\begin{figure*}[!t]
\centering
\includegraphics[width=\textwidth]{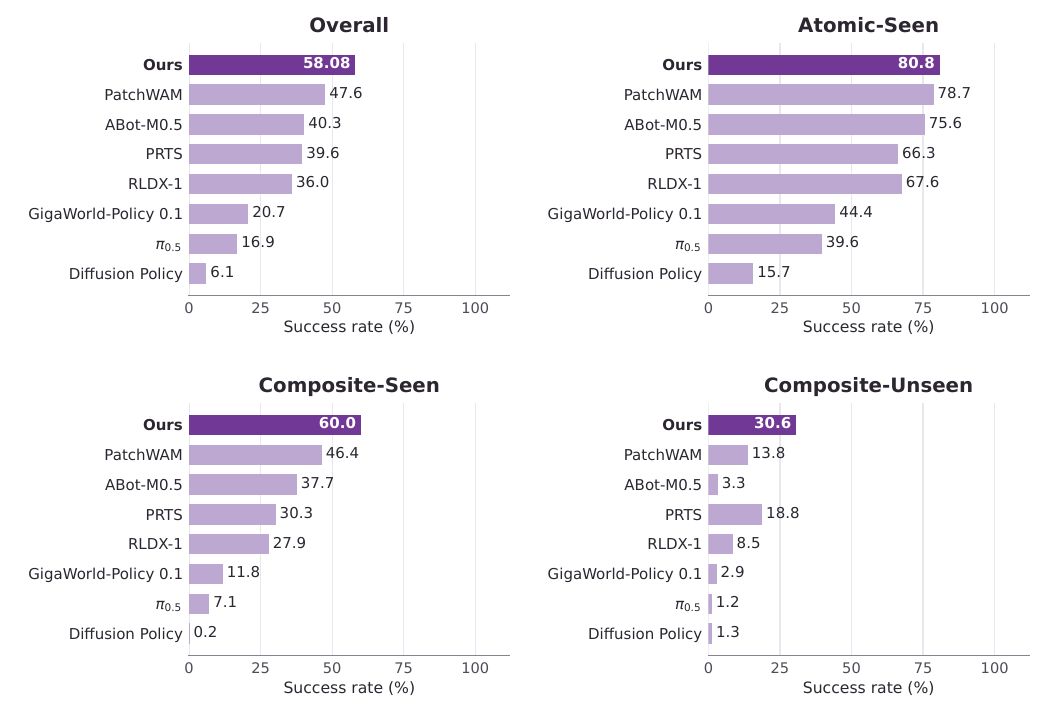}
\caption{Success rates (\%) on the 50 RoboCasa365 evaluation tasks, reported as the overall score and for the three splits. Baseline results are taken from the official leaderboard~\cite{robocasaLeaderboard2026}; the overall score is the average over all 50 tasks.}
\label{fig:robocasa365}
\end{figure*}

\paragraph{RoboCasa365.}
As shown in Fig.~\ref{fig:robocasa365}, \model achieves an overall success rate of $58.08\%$, with $80.8\%$, $60.0\%$, and $30.6\%$ on the Atomic-Seen, Composite-Seen, and Composite-Unseen splits, respectively.
Among the compared methods, PatchWAM obtains the highest overall score of $47.6\%$, which is $10.48$ percentage points lower than that of \model.

\paragraph{Real-robot tasks.}
As shown in Tab.~\ref{tab:real}, \model achieves a mean success rate of $78.8\%$ over the four real-robot tasks, compared with $65.0\%$ for the strongest baseline, $\pi_{0.5}$.
On block transfer and cap unscrewing, which are performed by two LeKiwi robots, \model reaches $88\%$ and $70\%$, respectively.
On starfruit bagging and block passing, which are performed by two LeKiwi robots and one OpenArm, it reaches $76\%$ and $81\%$, respectively.

\begin{table}[H]
\centering
\scriptsize
\setlength{\tabcolsep}{3pt}
\caption{Success rates (\%) on the real-robot tasks. Mean is the unweighted average over the four tasks.}
\label{tab:real}
\begin{tabular}{@{}lccccc@{}}
\toprule
Method & \shortstack{Block\\transfer} & \shortstack{Cap\\unscrewing} & \shortstack{Starfruit\\bagging} & \shortstack{Block\\passing} & Mean \\
\midrule
Diffusion Policy~\cite{chi2023diffusionpolicy} & 31 & 18 & 16 & 26 & 22.8 \\
Global GauDP~\cite{wang2025gaudp} & 44 & 25 & 29 & 33 & 32.8 \\
Local GauDP~\cite{wang2025gaudp} & 40 & 21 & 18 & 22 & 25.3 \\
$\pi_{0.5}$~\cite{black2025pi05} & 77 & 64 & 58 & 61 & 65.0 \\
ACT~\cite{zhao2023learning} & 58 & 33 & 41 & 53 & 46.3 \\
Fast-WAM~\cite{yuan2026fastwam} & 71 & 51 & 63 & 69 & 63.5 \\
\midrule
\textbf{Ours} & \textbf{88} & \textbf{70} & \textbf{76} & \textbf{81} & \textbf{78.8} \\
\bottomrule
\end{tabular}
\end{table}

\section{Conclusion}
\label{sec:concl}

\model places a shared world state $\phistate$ at the center of multi-robot action prediction.
Calibrated observations from available cameras are aggregated into this state.
Each robot query reads the same state, forming an all-to-one, one-to-all flow of visual information.
Vision--language context and relative-pose-aware query exchange support coordination, while the flow-based decoder retains a separate action stream for each robot.
Masked-view cooperative training withholds selected cameras from both visual branches while preserving action supervision for every robot, encouraging the use of observations shared by teammates.
\model achieves $74.0\%$ mean success on RoboFactory and $78.8\%$ across four real-robot tasks, and reaches $58.08\%$ overall on RoboCasa365.
The shared-state design thus provides a common spatial reference for the team while preserving robot-specific control.

{
    \small
    \bibliographystyle{ieeenat_fullname}
    \bibliography{main}

@inproceedings{brohan2023rt2,
  title={{RT-2}: Vision-Language-Action Models Transfer Web Knowledge to Robotic Control},
  author={Zitkovich, Brianna and Yu, Tianhe and Xu, Sichun and Xu, Peng and Xiao, Ted and Xia, Fei and Wu, Jialin and Wohlhart, Paul and Welker, Stefan and Wahid, Ayzaan and others},
  booktitle={Proceedings of the 7th Conference on Robot Learning},
  series={Proceedings of Machine Learning Research},
  volume={229},
  pages={2165--2183},
  publisher={PMLR},
  year={2023},
  url={https://proceedings.mlr.press/v229/zitkovich23a.html}
}

@inproceedings{kim2024openvla,
  title={{OpenVLA}: An Open-Source Vision-Language-Action Model},
  author={Kim, Moo Jin and Pertsch, Karl and Karamcheti, Siddharth and Xiao, Ted and Balakrishna, Ashwin and Nair, Suraj and Rafailov, Rafael and Foster, Ethan P and Sanketi, Pannag R and Vuong, Quan and Kollar, Thomas and Burchfiel, Benjamin and Tedrake, Russ and Sadigh, Dorsa and Levine, Sergey and Liang, Percy and Finn, Chelsea},
  booktitle={Proceedings of the 8th Conference on Robot Learning},
  series={Proceedings of Machine Learning Research},
  volume={270},
  pages={2679--2713},
  publisher={PMLR},
  year={2025},
  url={https://proceedings.mlr.press/v270/kim25c.html}
}

@inproceedings{black2024pi0,
  title={{$\pi_0$}: A Vision-Language-Action Flow Model for General Robot Control},
  author={Black, Kevin and Brown, Noah and Driess, Danny and Esmail, Adnan and Equi, Michael Robert and Finn, Chelsea and Fusai, Niccolo and Groom, Lachy and Hausman, Karol and Ichter, Brian and others},
  booktitle={Proceedings of Robotics: Science and Systems (RSS)},
  year={2025},
  doi={10.15607/RSS.2025.XXI.010}
}

@inproceedings{black2025pi05,
  title={{$\pi_{0.5}$}: a Vision-Language-Action Model with Open-World Generalization},
  author={Black, Kevin and Brown, Noah and Darpinian, James and Dhabalia, Karan and Driess, Danny and Esmail, Adnan and Equi, Michael Robert and Finn, Chelsea and Fusai, Niccolo and others},
  booktitle={Proceedings of the 9th Conference on Robot Learning},
  series={Proceedings of Machine Learning Research},
  volume={305},
  pages={17--40},
  publisher={PMLR},
  year={2025},
  url={https://proceedings.mlr.press/v305/black25a.html}
}

@article{bjorck2025groot,
  title={{GR00T N1}: An Open Foundation Model for Generalist Humanoid Robots},
  author={{NVIDIA} and Bjorck, Johan and Casta{\~n}eda, Fernando and Cherniadev, Nikita and Da, Xingye and Ding, Runyu and Fan, Linxi and Fang, Yu and Fox, Dieter and Hu, Fengyuan and others},
  journal={arXiv preprint arXiv:2503.14734},
  year={2025}
}

@inproceedings{du2023universal,
  title={Learning Universal Policies via Text-Guided Video Generation},
  author={Du, Yilun and Yang, Sherry and Dai, Bo and Dai, Hanjun and Nachum, Ofir and Tenenbaum, Josh and Schuurmans, Dale and Abbeel, Pieter},
  booktitle={Advances in Neural Information Processing Systems (NeurIPS)},
  year={2023}
}

@inproceedings{li2025uva,
  title={Unified Video Action Model},
  author={Li, Shuang and Gao, Yihuai and Sadigh, Dorsa and Song, Shuran},
  booktitle={Proceedings of Robotics: Science and Systems (RSS)},
  year={2025},
  address={Los Angeles, CA, USA},
  month={June},
  doi={10.15607/RSS.2025.XXI.074}
}

@inproceedings{zhu2025uwm,
  title={Unified World Models: Coupling Video and Action Diffusion for Pretraining on Large Robotic Datasets},
  author={Zhu, Chuning and Yu, Raymond and Feng, Siyuan and Burchfiel, Benjamin and Shah, Paarth and Gupta, Abhishek},
  booktitle={Proceedings of Robotics: Science and Systems (RSS)},
  year={2025},
  address={Los Angeles, CA, USA},
  month={June},
  doi={10.15607/RSS.2025.XXI.015}
}

@inproceedings{qin2025robofactory,
  title={{RoboFactory}: Exploring Embodied Agent Collaboration with Compositional Constraints},
  author={Qin, Yiran and Kang, Li and Song, Xiufeng and Yin, Zhenfei and Liu, Xiaohong and Liu, Xihui and Zhang, Ruimao and Bai, Lei},
  booktitle={Proceedings of the IEEE/CVF International Conference on Computer Vision (ICCV)},
  pages={10075--10085},
  year={2025}
}

@inproceedings{ze2024dp3,
  title={{3D Diffusion Policy}: Generalizable Visuomotor Policy Learning via Simple {3D} Representations},
  author={Ze, Yanjie and Zhang, Gu and Zhang, Kangning and Hu, Chenyuan and Wang, Muhan and Xu, Huazhe},
  booktitle={Proceedings of Robotics: Science and Systems (RSS)},
  year={2024},
  doi={10.15607/RSS.2024.XX.067}
}

@inproceedings{wang2025gaudp,
  title={{GauDP}: Reinventing Multi-Agent Collaboration through Gaussian-Image Synergy in Diffusion Policies},
  author={Wang, Ziye and Kang, Li and Qin, Yiran and Ma, Jiahua and Peng, Zhanglin and Bai, Lei and Zhang, Ruimao},
  booktitle={Advances in Neural Information Processing Systems (NeurIPS)},
  year={2025},
  url={https://proceedings.nips.cc/paper_files/paper/2025/hash/0895de150660b6c2aef9b0f098cb1dcd-Abstract-Conference.html}
}

@article{park2026clsdp,
  title={Distilling Collaborative Dynamics into Latent Space for Implicit Coordination in Decentralized Multi-Agent Manipulation},
  author={Park, Chanyoung and Yoon, Minsung and Jeong, Andrew and Yoon, Sung-eui},
  journal={arXiv preprint arXiv:2606.22982},
  year={2026},
  url={https://arxiv.org/abs/2606.22982}
}

@article{yuan2026fastwam,
  title={{Fast-WAM}: Do World Action Models Need Test-Time Future Imagination?},
  author={Yuan, Tianyuan and Dong, Zibin and Liu, Yicheng and Zhao, Hang},
  journal={arXiv preprint arXiv:2603.16666},
  year={2026},
  url={https://arxiv.org/abs/2603.16666}
}

@inproceedings{nasiriany2024robocasa,
  title={{RoboCasa}: Large-Scale Simulation of Everyday Tasks for Generalist Robots},
  author={Nasiriany, Soroush and Maddukuri, Abhiram and Zhang, Lance and Parikh, Adeet and Lo, Aaron and Joshi, Abhishek and Mandlekar, Ajay and Zhu, Yuke},
  booktitle={Proceedings of Robotics: Science and Systems (RSS)},
  year={2024},
  doi={10.15607/RSS.2024.XX.050}
}

@inproceedings{nasiriany2026robocasa365,
  title={{RoboCasa365}: A Large-Scale Simulation Framework for Training and Benchmarking Generalist Robots},
  author={Nasiriany, Soroush and Nasiriany, Sepehr and Maddukuri, Abhiram and Zhu, Yuke},
  booktitle={International Conference on Learning Representations (ICLR)},
  year={2026},
  note={arXiv:2603.04356}
}

@misc{robocasaLeaderboard2026,
  title={{RoboCasa 365 Leaderboard}},
  author={{RoboCasa Team}},
  year={2026},
  howpublished={\url{https://robocasa.ai/leaderboard.html}},
  note={Updated October 19, 2026; accessed October 21, 2026}
}

@article{wang2026patchwam,
  title={An Action Is Worth One Patch: Unified World-Action Modeling with {PatchWAM}},
  author={Wang, Tianheng and Xie, Zhou and Jia, Heng and Xu, Jianhua and Zhang, Tong and Yu, Kaicheng},
  journal={arXiv preprint arXiv:2609.25961},
  year={2026}
}

@article{chen2026abotm05,
  title={{ABot-M0.5}: Unified Mobility-and-Manipulation World Action Model},
  author={Chen, Ronghan and Yang, Yandan and Tang, Zuojin and Huo, Dongjie and Lin, Tong and Wu, Haoning and Liu, Haoyun and Chen, Yuzhi and Zheng, Lulu and Yuan, Botai and others},
  journal={arXiv preprint arXiv:2607.00678},
  year={2026}
}

@article{zhang2026prts,
  title={{PRTS}: A Primitive Reasoning and Tasking System via Contrastive Representations},
  author={Zhang, Yang and Zhao, Jiangyuan and Fan, Chenyou and Yan, Fangzheng and Li, Tian and Tang, Haitong and Fu, Sen and Wu, Xuan'er and Weng, Qizhen and Zhang, Weinan and others},
  journal={arXiv preprint arXiv:2604.27472},
  year={2026}
}

@article{kim2026rldx1,
  title={{RLDX-1} Technical Report},
  author={Kim, Dongyoung and Jang, Huiwon and Koo, Myungkyu and Jang, Suhyeok and Kim, Taeyoung and others},
  journal={arXiv preprint arXiv:2605.03269},
  year={2026}
}

@article{ye2026gigaworldpolicy,
  title={{GigaWorld-Policy}: An Efficient Action-Centered World--Action Model},
  author={Ye, Angen and Wang, Boyuan and Ni, Chaojun and Huang, Guan and Zhao, Guosheng and Li, Hao and Li, Hengtao and others},
  journal={arXiv preprint arXiv:2603.17240},
  year={2026}
}

@article{bai2025qwen3vl,
  title={{Qwen3-VL} Technical Report},
  author={Bai, Shuai and Cai, Yuxuan and Chen, Ruizhe and Chen, Keqin and Chen, Xionghui and Cheng, Zesen and Deng, Lianghao and Ding, Wei and Gao, Chang and Ge, Chunjiang and others},
  journal={arXiv preprint arXiv:2511.21631},
  year={2025}
}

@inproceedings{chan2022eg3d,
  title={Efficient Geometry-Aware {3D} Generative Adversarial Networks},
  author={Chan, Eric R. and Lin, Connor Z. and Chan, Matthew A. and Nagano, Koki and Pan, Boxiao and De Mello, Shalini and Gallo, Orazio and Guibas, Leonidas J. and Tremblay, Jonathan and Khamis, Sameh and Karras, Tero and Wetzstein, Gordon},
  booktitle={Proceedings of the IEEE/CVF Conference on Computer Vision and Pattern Recognition (CVPR)},
  pages={16123--16133},
  year={2022}
}

@inproceedings{tian2025pdfactor,
  title={{PDFactor}: Learning Tri-Perspective View Policy Diffusion Field for Multi-Task Robotic Manipulation},
  author={Tian, Jingyi and Wang, Le and Zhou, Sanping and Wang, Sen and Li, Jiayi and Sun, Haowen and Tang, Wei},
  booktitle={Proceedings of the IEEE/CVF Conference on Computer Vision and Pattern Recognition (CVPR)},
  pages={15757--15767},
  year={2025}
}

@inproceedings{liu2022rectified,
  title={Flow Straight and Fast: Learning to Generate and Transfer Data with Rectified Flow},
  author={Liu, Xingchao and Gong, Chengyue and Liu, Qiang},
  booktitle={International Conference on Learning Representations (ICLR)},
  year={2023},
  note={arXiv:2209.03003}
}

@article{chi2023diffusionpolicy,
  title={Diffusion Policy: Visuomotor Policy Learning via Action Diffusion},
  author={Chi, Cheng and Xu, Zhenjia and Feng, Siyuan and Cousineau, Eric and Du, Yilun and Burchfiel, Benjamin and Tedrake, Russ and Song, Shuran},
  journal={The International Journal of Robotics Research},
  volume={44},
  number={10--11},
  pages={1684--1704},
  year={2025},
  doi={10.1177/02783649241273668}
}

@inproceedings{team2024octo,
  title={Octo: An Open-Source Generalist Robot Policy},
  author={Ghosh, Dibya and Walke, Homer Rich and Pertsch, Karl and Black, Kevin and Mees, Oier and Dasari, Sudeep and Hejna, Joey and Kreiman, Tobias and Xu, Charles and Luo, Jianlan and others},
  booktitle={Proceedings of Robotics: Science and Systems (RSS)},
  year={2024},
  doi={10.15607/RSS.2024.XX.090}
}

@inproceedings{zhao2023learning,
  title={Learning Fine-Grained Bimanual Manipulation with Low-Cost Hardware},
  author={Zhao, Tony Z. and Kumar, Vikash and Levine, Sergey and Finn, Chelsea},
  booktitle={Robotics: Science and Systems (RSS)},
  year={2023}
}

@inproceedings{zhou2019continuity,
  title={On the Continuity of Rotation Representations in Neural Networks},
  author={Zhou, Yi and Barnes, Connelly and Lu, Jingwan and Yang, Jimei and Li, Hao},
  booktitle={Proceedings of the IEEE/CVF Conference on Computer Vision and Pattern Recognition (CVPR)},
  pages={5745--5753},
  year={2019}
}

@misc{lekiwi2025,
  title={{LeKiwi}: Low-Cost Mobile Manipulator},
  author={{SIGRobotics-UIUC}},
  year={2025},
  howpublished={\url{https://github.com/SIGRobotics-UIUC/LeKiwi}},
  note={Accessed October 26, 2026}
}

@misc{openarm2025,
  title={{OpenArm}: A Fully Open-Source Humanoid Arm for Physical {AI} Research and Deployment in Contact-Rich Environments},
  author={{Enactic}},
  year={2025},
  howpublished={\url{https://github.com/enactic/openarm}},
  note={Accessed October 26, 2026}
}
}
\end{document}